\documentclass[letterpaper, 10pt, conference]{ieeeconf}      

\IEEEoverridecommandlockouts                              

\usepackage[table]{xcolor}
\definecolor{bedColor}{rgb}{0, 0, 1}
\definecolor{booksColor}{rgb}{0.9137,0.3490,0.1882}
\definecolor{ceilColor}{rgb}{0, 0.8549, 0}
\definecolor{chairColor}{rgb}{0.5843,0,0.9412}
\definecolor{floorColor}{rgb}{0.8706,0.9451,0.0941}
\definecolor{furnColor}{rgb}{1.0000,0.8078,0.8078}
\definecolor{objsColor}{rgb}{0,0.8784,0.8980}
\definecolor{paintColor}{rgb}{0.4157,0.5333,0.8000}
\definecolor{sofaColor}{rgb}{0.4588,0.1137,0.1608}
\definecolor{tableColor}{rgb}{0.9412,0.1373,0.9216}
\definecolor{tvColor}{rgb}{0,0.6549,0.6118}
\definecolor{wallColor}{rgb}{0.9765,0.5451,0}
\definecolor{windColor}{rgb}{0.8824,0.8980,0.7608}

\definecolor{buildingColor}{rgb}{0.5020, 0, 0}
\definecolor{vegetationColor}{rgb}{0.5020, 0.5020, 0}
\definecolor{carColor}{rgb}{0.2510, 0, 0.5020}
\definecolor{roadColor}{rgb}{0.5020, 0.2510, 0.5020}
\definecolor{fenceColor}{rgb}{0.2510, 0.7529, 0}
\definecolor{sidewalkColor}{rgb}{0, 0, 0.7529}
\definecolor{poleColor}{rgb}{0, 0.2510, 0.2510}

\usepackage{stmaryrd}
\usepackage{algorithm, algorithmicx, algpseudocode}
\usepackage{multirow}
\usepackage{caption}
\usepackage{graphicx}
\graphicspath{{./}}

\usepackage{amsmath,amssymb,thmtools,enumerate}
\usepackage[noadjust]{cite}
\usepackage{setspace}
\usepackage{booktabs}
\usepackage[caption=false,font=footnotesize]{subfig}

\usepackage{soul} 

\newcommand{\Ccal}{\mathcal{C}}
\newcommand{\Dcal}{\mathcal{D}}

\newcommand{\Mcal}{\mathcal{M}}

\newcommand{\Ocal}{\mathcal{O}}

\newcommand{\Xcal}{\mathcal{X}}

\newcommand{\Zcal}{\mathcal{Z}}

\newcommand\crule[3][black]{\textcolor{#1}{\rule{#2}{#3}}}
\newcommand{\diag}{\mathop{\mathrm{diag}}}

\newcommand{\bigO}[1]{\Ocal(#1)} 

\newcommand{\squeezeup}{\vspace{-3mm}}
\DeclareMathOperator*{\argmax}{arg\,max}  
\newcommand{\transpose}{\mathsf{T}}

\makeatletter
\algrenewcommand\ALG@beginalgorithmic{\small}
\algrenewcommand\algorithmiccomment[2][\small]{{#1\(\sslash\) #2}}
\makeatother

\title{\LARGE \bf Sparse Bayesian Inference for Dense Semantic Mapping}

\author{Lu Gan, Maani Ghaffari Jadidi, Steven A. Parkison, and Ryan M. Eustice%
\thanks{This work was supported by a grant from the Toyota Research Institute under award N021515.} \\%
\thanks{The authors are with College of Engineering, \mbox{University of Michigan}, Ann Arbor, MI 48109 USA {\tt\small \{{ganlu, maanigj, sparki, eustice\}}@umich.edu}}%
}

\begin{document}
\maketitle
\thispagestyle{empty}
\pagestyle{empty}

\begin{abstract}
Despite impressive advances in simultaneous localization and mapping, dense robotic mapping remains challenging due to its inherent nature of being a high-dimensional inference problem. In this paper, we propose a dense semantic robotic mapping technique that exploits sparse Bayesian models, in particular, the relevance vector machine, for high-dimensional sequential inference. The technique is based on the principle of automatic relevance determination and produces sparse models that use a small subset of the original dense training set as the dominant basis. The resulting map posterior is continuous, and queries can be made efficiently at any resolution. Moreover, the technique has probabilistic outputs per semantic class through Bayesian inference. We evaluate the proposed relevance vector semantic map using publicly available benchmark datasets, NYU Depth V2 and KITTI; and the results show promising improvements over the state-of-the-art techniques.
\end{abstract}


\IEEEpeerreviewmaketitle

\section{INTRODUCTION}
\begin{figure}[t]
\centering 
  \subfloat{\includegraphics[width=.99\columnwidth,trim={0cm 0cm 0cm 0cm},clip]{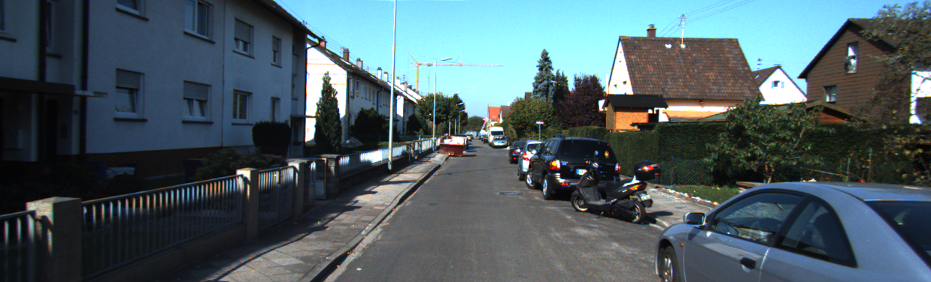}
  \label{fig:kittiimg}}\\ \squeezeup
  \subfloat{\includegraphics[width=.99\columnwidth,trim={0cm 0cm 0cm 0cm},clip]{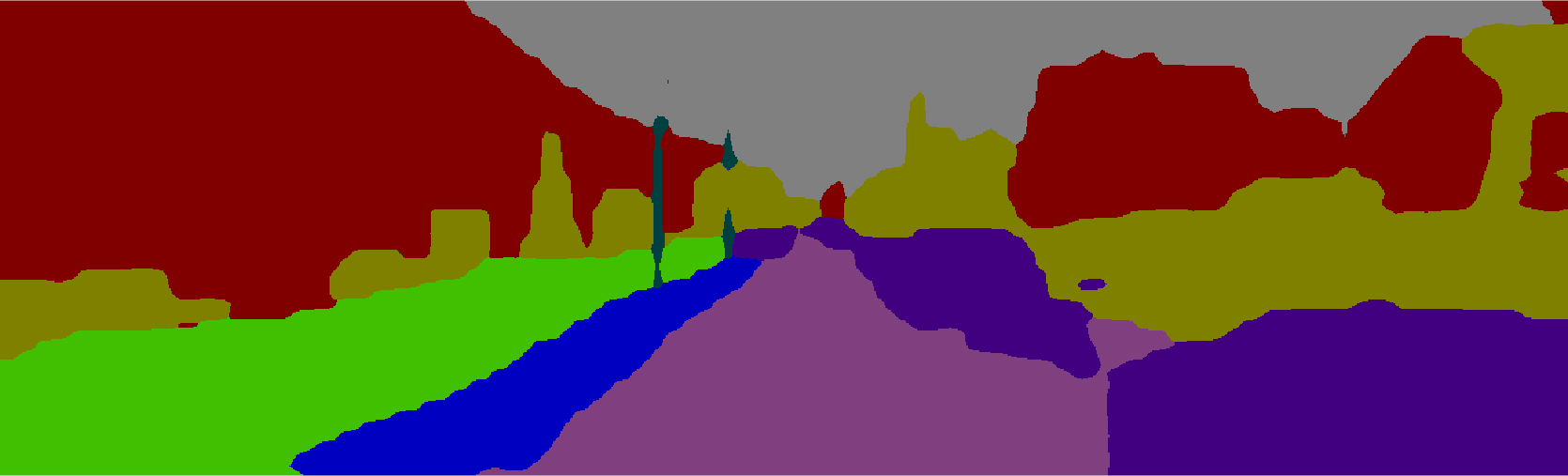}
  \label{fig:gtseg}} \\ \squeezeup
  \subfloat{\includegraphics[width=.99\columnwidth,trim={0cm 0cm 2.5cm 0cm},clip]{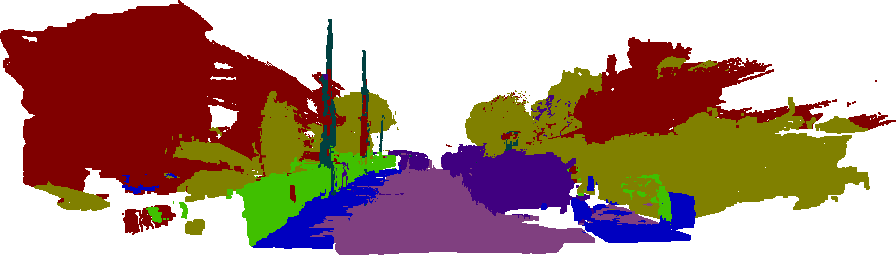}
\label{fig:kittirvsm}} \\ 
\subfloat{\crule[buildingColor]{0.4cm}{0.3cm} \scriptsize{Building}
\crule[vegetationColor]{0.4cm}{0.3cm} \scriptsize{Vegetation}
\crule[carColor]{0.4cm}{0.3cm} \scriptsize{Car}
\crule[roadColor]{0.4cm}{0.3cm} \scriptsize{Road}
\crule[fenceColor]{0.4cm}{0.3cm} \scriptsize{Fence}
\crule[sidewalkColor]{0.4cm}{0.3cm} \scriptsize{Sidewalk}
\crule[poleColor]{0.4cm}{0.3cm} \scriptsize{Pole}
}
\caption{An example of the RVSM using KITTI odometry sequence 15~\cite{Geiger2012CVPR}. The input image is shown in the top. The middle figure shows the 2D semantic segmentation of the scene. The bottom figure shows the corresponding 3D dense RVSM with the classes given in the legend below.}
\label{fig:first}
\squeezeup\squeezeup
\end{figure}

Dense mapping is a complex and high-dimensional inference problem in robotic perception~\cite{Thrun:2003:RMS:779343.779345}. Traditionally, dense maps are built using occupancy maps~\cite{hornung2013octomap,merali2014optimizing}, which have become successful in robotic exploration and navigation  tasks~\cite{charrow2015information}. The incorporation of the semantic knowledge into robotic perception systems have enriched the representation of the environment and improved scene understanding~\cite{bowman2017probabilistic,runz2017co} (e.g., see Fig.~\ref{fig:first}). Advances in \emph{deep convolutional neural networks} for semantic segmentation provide reasonably high performance in both indoor and outdoor benchmarks together with promising processing time for pixel-wise estimation~\cite{eigen2015predicting,YuKoltun2016}. The availability of high performance and fast image segmentation techniques has led to a series of works on three-dimensional (3D) dense semantic map building with promising outcomes~\cite{valentin2013mesh,sengupta2013urban,ladicky2014associative,vineet2015incremental,sengupta2015semantic,wolf2015fast,mccormac2017semanticfusion,ma2017multi,yang2017semantic}.

The most common approach to build a semantic map is using a voxelized representation of 3D space and \emph{undirected graphical models} (such as \emph{Markov random fields}) or \emph{Conditional Random Fields} (CRFs) to model the conditional independence between voxels~\cite{murphy2012machine}. CRFs are \emph{discriminative} models that are suitable for spatial inference. However, the main drawback of the approach mentioned earlier is that the map representation is discrete from the beginning. Thus, the map resolution is often fixed or limited and once the map is inferred the resolution cannot be increased. Instead, we argue that it is beneficial to model a continuous model of the semantic map, and discretize the representation when it is required, for example, for fast map fusion or a particular application such as robotic navigation. In~\cite{paul2012semantic,gpsm2017rssws}, \emph{Gaussian processes} (GPs) semantic map representation is proposed, which is inherently continuous, and prediction can be made at any desired location. The semantic mapping is formulated as a \emph{multi-class classification} problem and raw pixelated semantic measurements (class labels) are used to generalize the traditional occupied and unoccupied class assignment. GPs provide a fully Bayesian and probabilistic framework for high-dimensional inference problems such as dense robotic mapping. However, the GP training and query time complexities are cubic in the number of \emph{training points} and quadratic in the number \emph{query points}, respectively, which are limiting factors.

In this paper, we propose a dense semantic robotic mapping technique that exploits \emph{sparse Bayesian models}, in particular, the \emph{Relevance Vector Machine} (RVM), for high-dimensional sequential inference~\cite{tipping2001sparse,tipping2003fast}. RVM has many nice properties of GPs such as continuity, Bayesian inference, and probabilistic outputs, and is sparse through using the principle of \emph{Automatic Relevance Determination} (ARD)~\cite{neal1996bayesian}. In addition, RVM allows for fast and efficient queries at any desired location. We build a 3D semantic map representation called \emph{Relevance Vector Semantic Map} (RVSM). The present work has the following contributions:
\begin{itemize}
\item The proposed technique can infer missing labels and deal with sparse measurements.
\item RVSM is continuous and sparse, and queries can be made efficiently at any desired locations; therefore, the map can be inferred with any resolution.
\item It generalizes occupancy maps, i.e.\@ binary maps, to the multi-class semantic representation, which provides rich maps for robotic planning tasks.
\item We evaluate the performance of RVSM using benchmark datasets and compare it with the available state-of-the-art techniques.
\end{itemize}

\textbf{Outline.} In the following section, a review of the related work is given. Sparse Bayesian models and RVMs are explained in Section~\ref{sec:sparsebayes}. The detailed formulation of relevance vector semantic mapping is presented in Section~\ref{sec:rvsm}. Section~\ref{sec:complexity} includes the time complexity analysis of the proposed relevance vector semantic map. The comparison of mapping results using two publicly available datasets are presented in Section~\ref{sec:results}, as well as discussions on the limitations of the proposed technique; and finally, Section~\ref{sec:conclusion} concludes the paper and discusses possible extensions of this work.

\textbf{Notation.} The probabilities and probability densities are not distinguished in general. Matrices are capitalized in bold, such as in $\boldsymbol X$, and vectors are in lower case bold type, such as in $\boldsymbol x$. Vectors are column-wise and $1\colon n$ means integers from $1$ to $n$. The Euclidean norm is shown by $\lVert \cdot \rVert$. Random variables, such as $X$, and their realizations, $x$, are sometimes denoted interchangeably where it is evident from context. An alphabet such as $\mathcal{X}$ denotes a set. A subscript asterisk, such as in $\boldsymbol x_*$, indicates a reference to a test set quantity. $\mathrm{vec}(x_1,\dots,x_n)$ denotes a vector such as $\boldsymbol x$ constructed by stacking $x_i$, $\forall \ i \in \{1\colon n\}$.

\section{Related Work}
\label{sec:related}
Early works in dense semantic mapping back-project labels from a segmented image to the reconstructed 3D points, and assign each voxel or mesh face to the most frequent label according to a label histogram~\cite{he2013nonparametric,sengupta2013urban}. Bayesian frameworks are also utilized to fuse labels from multiple views into a voxel-based 3D map. In~\cite{stuckler2012semantic}, probabilistic segmentation outputs of multiple images obtained by \emph{Random decision Forests} (RFs) are transfered into 3D and updated using a \emph{Bayes filter}. In~\cite{xiang2017darnn}, DA-RNN is proposed, which integrates a deep network for RGB-D video labeling into a dense 3D reconstruction framework built by KinectFusion. DA-RNN yields consist semantic labeling of indoor 3D scenes; however, it is assumed that semantic labels and geometric information are independent, and therefore, the consistency largely depends on the performance of data association computed by KinectFusion. The 3D label fusion is done by updating a probability vector of semantic classes and choosing the label with the maximum probability. 

In~\cite{kim20133d}, a Voxel-CRF model is proposed to capture the geometric and semantic relationships by constructing a CRF over  the 3D volume. A CRF-over-mesh model is also proposed for semantic modeling of both indoor and outdoor scenes~\cite{valentin2013mesh}. In~\cite{hermans2014dense}, a Kalman filter is used to transfer 2D class probabilities obtained by RFs to the 3D model, and 3D labels are further refined through a dense pairwise CRF over the point cloud. In~\cite{wolf2015fast} a similar RFs-CRFs framework is used together with an efficient mean-field CRF inference method to speed up the mapping process. In~\cite{zhao2016building}, a higher-order CRF model is used to enforce temporal label consistency by generating higher-order \emph{cliques} from \emph{superpixels} correspondences in an RGB-D video, which improved the precision of semantic maps. In SemanticFusion~\cite{mccormac2017semanticfusion}, a fully-connected CRF with Gaussian \emph{edge potentials} is applied to \emph{surfels} by incrementally updating probability distributions. 

Semantic mapping for outdoor scenes is important for robot applications such as autonomous driving. In~\cite{kundu2014joint}, a 3D semantic reconstruction approach for outdoor scenes using 3D CRFs is proposed. However, to achieve large-scale dense 3D maps, memory and computational efficiency can cause a bottleneck. In~\cite{vineet2015incremental}, the memory-friendly hash-based 3D volumetric representation and a CRF are used for incremental dense semantic mapping of large scenes with a near real-time processing time. Semantic Octree~\cite{sengupta2015semantic} constructs a higher-order CRF model over voxels in OctoMap~\cite{hornung2013octomap} to a multi-resolution 3D semantic map representation; higher-oder cliques are naturally defined as internal nodes in the hierarchical octree data structure. In~\cite{yang2017semantic} a similar CRF model together with 3D scrolling occupancy grids are proposed and the reported results are using KITTI dataset~\cite{Geiger2012CVPR} are promising.

A common feature of the works mentioned earlier is the discretization of the space prior to map inference which means, once the map is inferred, the prediction cannot be computed at any arbitrary points. In this paper, we propose a novel and alternative solution for the problem of dense 3D map building that is continuous and at the same time sparse. RVSM incrementally learns relevance vectors which are dominant basis in the data, and builds a sparse Bayesian model of the 3D map. As a result, the prediction can be made efficiently at any desired location. The training process that is often more expensive is accelerated by utilizing a sequential sparse Bayesian learning algorithm in~\cite{tipping2003fast}. We evaluate RVSM using NYU Depth V2 (NYUDv2)~\cite{Silberman:ECCV12} and KITTI datasets and compare the achieved results with the recent works mentioned here.

\section{Sparse Bayesian Modeling and Relevance Vector Machine}
\label{sec:sparsebayes}
A sparse linear model (linear in weights) takes the following form~\cite{tipping2004bayesian,bishop2006pattern}:
\begin{equation}
\label{eq:sbm_targets}
y(\boldsymbol x ; \boldsymbol w) \triangleq \sum_{j=1}^{n_b} w_j \phi_j (\boldsymbol x) = \boldsymbol w^{\transpose} \boldsymbol \phi (\boldsymbol x)
\end{equation}
where $y(\boldsymbol x ; \boldsymbol w)$ is an approximation to the \emph{latent} function of interest, which is real-valued for regression and discriminant for classification problems; $\boldsymbol x \triangleq \mathrm{vec}(x_1, \dots, x_{d})$ is the \mbox{$d$-dimensional} input vector, and $\boldsymbol w \triangleq \mathrm{vec}(w_1, \dots, w_{n_b}) $ is the parameter vector (weights) of a set of $n_b$ basis vectors $\boldsymbol \phi(\boldsymbol x) \triangleq \mathrm{vec}(\phi_1(\boldsymbol x), \dots, \phi_{n_b}(\boldsymbol x))$. Let $\boldsymbol t \triangleq \mathrm{vec}(t_1,\dots,t_{n_t})$ be the \emph{target} (or output) vector and a set of input-target pairs $\Dcal \triangleq \{\boldsymbol x_i, t_i\}^{n_t}_{i=1}$ be the training set where $n_t$ is the number of examples or measurements and $t_i$ is the noisy measurement of the latent $y_i$. The objective is to learn or infer the parameter vector $\boldsymbol w$ of the function $y(\boldsymbol x; \boldsymbol w)$ such that it generalizes well to new inputs $\boldsymbol x_*$ (test data). 
 
Instead of computing a \emph{point estimate} for the weights, a Bayesian predictive framework infers the \emph{posterior distribution} over $\boldsymbol w$ via Bayes' rule:
\begin{equation}
\label{eq:sbm_posterior}
p(\boldsymbol w | \boldsymbol t, \boldsymbol \alpha) = \frac{p(\boldsymbol t | \boldsymbol w) p(\boldsymbol w | \boldsymbol \alpha)}{p(\boldsymbol t | \boldsymbol \alpha)}
\end{equation}
where $p(\boldsymbol t | \boldsymbol w)$ is the \emph{likelihood} of training data, \mbox{$p(\boldsymbol t | \boldsymbol \alpha) = \int p(\boldsymbol t | \boldsymbol w) p(\boldsymbol w | \boldsymbol \alpha) \mathrm{d} \boldsymbol w$} is the \emph{marginal likelihood}, and $p(\boldsymbol w | \boldsymbol \alpha)$ is the \emph{prior distribution} defined over $\boldsymbol w$ using the principle of ARD to set the inverse variance \emph{hyperparameter} as $\boldsymbol \alpha \triangleq \mathrm{vec}(\alpha_1, \dots, \alpha_{n_b})$. By placing a zero-mean Gaussian prior over the weights we have
\begin{equation}
 \label{eq:wprior}
 p(\boldsymbol w | \boldsymbol \alpha) = (2\pi)^{-n_b/2} \prod_{j=1}^{n_b} \alpha_j^{1/2} \exp(-\frac{\alpha_j w_j^2}{2})
\end{equation}
where each $\alpha_j$ is individually controlling the effect of the prior over its associated weight. In fact, this form of prior using ARD ultimately makes the model sparse. As many of the hyperparameters approach infinity during inference, the posterior distributions of their associated weights are peaked around zero. Consequently, only a few basis functions with non-zero weights survive in the final model, resulting in a sparse model. The basis functions that survive are called \emph{relevance vectors}.

In a truly Bayesian framework, a \emph{hyperprior} $p(\boldsymbol \alpha)$ should be placed to integrate out $\boldsymbol \alpha$ in the final predictive distribution; however, in practice, this approach is usually not analytically tractable. Hence, it is common to maximize the marginal likelihood to get a most-probable point estimate of $\boldsymbol \alpha$ from data:
\begin{equation}
\label{eq:sbm_marginal}
\boldsymbol \alpha^{\star}  = \argmax_{\boldsymbol \alpha} \ \log p(\boldsymbol t | \boldsymbol \alpha) 
\end{equation}
Once $\boldsymbol \alpha^{\star}$ is obtained, the predictive distribution of the target $t_*$ of a test data $x_*$ can be computed by marginalizing over the uncertain variables $\boldsymbol w$~\cite{tipping2001sparse}:
\begin{equation}
\label{eq:sbm_predict}
p(t_* | \boldsymbol t, \alpha^{\star}) = \int p(t_* | \boldsymbol w)p(\boldsymbol w | \boldsymbol t, \boldsymbol \alpha^{\star}) \mathrm{d} \boldsymbol w
\end{equation}

To build a binary classifier, we adopt a Bernoulli likelihood as
\begin{equation}
\label{eq:rvm_classification}
p(\boldsymbol t | \boldsymbol w) = \prod_{i = 1}^{n_t} \sigma(y(\boldsymbol x_i ; \boldsymbol w))^{t_i}
\big[ 1 - \sigma(y( \boldsymbol x_i ; \boldsymbol w)) ]^{1-t_i}
\end{equation}
where $\sigma(y) = 1/(1 + e^y)$ is the logistic sigmoid function, and the targets $t_i \in \{0, 1\}$. Unlike the regression case using a Gaussian likelihood, this likelihood formulation prevents the closed-form solutions since the weights in~\eqref{eq:sbm_predict} cannot be integrated out analytically; therefore, approximate inference strategies are required.
For the current $\boldsymbol \alpha^{\star}$, the most-probable weights $\boldsymbol w^{\star}$ can be estimated iteratively:
\begin{equation}
\label{eq:}
\begin{split}
&\boldsymbol w^{\star} = \argmax_{\boldsymbol w} \ \log p(\boldsymbol w | \boldsymbol t, \boldsymbol \alpha^{\star}) \\
& = \argmax_{\boldsymbol w} \ \log p(\boldsymbol t | \boldsymbol w) + \log p(\boldsymbol w | \boldsymbol \alpha^{\star}) \\
& = \argmax_{\boldsymbol w} \sum_{i=1}^{n_t} [t_i \log y_i + (1-t_i) \log(1-y_i)] - \frac{1}{2} \boldsymbol w^\transpose \boldsymbol A \boldsymbol w
\end{split}
\end{equation}
where $y_i \triangleq \sigma(y(\boldsymbol x_i;\boldsymbol w))$ and $\boldsymbol A \triangleq \diag(\alpha_1, \dots, \alpha_{n_b})$.

The Laplace approximation method provides a locally Gaussian approximation of the weight posterior around $\boldsymbol w^{\star} = \boldsymbol \mu$ with covariance $\boldsymbol \Sigma$:
\begin{equation}
\label{eq:fast_mu}
\boldsymbol \mu = \boldsymbol A^{-1} \boldsymbol \Phi^\transpose (\boldsymbol t - \boldsymbol y)
\end{equation}
\begin{equation}
\label{eq:fast_sigma}
\boldsymbol \Sigma = (\boldsymbol \Phi^\transpose \boldsymbol B \boldsymbol \Phi + \boldsymbol A)^{-1}
\end{equation}
where $\boldsymbol \Phi \triangleq [\boldsymbol \phi_1, \dots, \boldsymbol \phi_{n_b}]^\transpose$, $\boldsymbol B \triangleq \diag(\beta_1, \dots, \beta_{n_t}$) with \mbox{$\beta_i \triangleq \sigma(y_i)[1 - \sigma(y_i)]$}, and \mbox{$\boldsymbol y \triangleq \mathrm{vec}(y_1, \dots, y_{n_t})$}. 

The RVM is a specialization of a sparse Bayesian model using \emph{kernel} basis. The basis vector $\phi_j(\boldsymbol x)$ is replaced with a vector of kernel functions $\mathrm{vec}(1, k(\boldsymbol x_1 , \boldsymbol x_j) \dots, k(\boldsymbol x_{n_b} , \boldsymbol x_j))$, where $k(\cdot,\cdot)$ evaluates the similarity between two inputs by implicitly mapping them to a \emph{feature space}~\cite{scholkopf2002learning}. In the next section, we formulate the relevance vector semantic map using the binary RVM classifier and introduce the sequential sparse Bayesian learning algorithm for fast and incremental training.

\section{Relevance Vector Semantic Map}
\label{sec:rvsm}
In this section, we formulate the RVSM and introduce the sequential sparse Bayesian learning algorithm to infer a dense semantic map incrementally. Let $\Xcal \subset \mathbb{R}^3$ be the set of spatial coordinates to build a map on, and $\Ccal=\{c_k\}_{k=1}^{n_c}$ be the set of semantic class labels. Let \mbox{$\Zcal \subset \Xcal \times \Ccal$} be the set of possible measurements. The observation consists of an $n_z$-tuple random variable $(Z_1,\dots,Z_{n_z})$ whose elements can take values \mbox{$\boldsymbol z_i \in \Zcal$, $i \in \{1\colon n_z\}$} where $\boldsymbol z_i = \mathrm{vec}(\boldsymbol x_i,t_i)$, $\boldsymbol x_i \in \Xcal$, and $t_i \in \Ccal$. The training set is defined as $\Dcal \subseteq \Zcal$ where $n_t \leq n_z$ is the number of training points. Let $\Mcal$ be the set of possible semantic maps. We consider the map of the environment as an $n_m$-tuple random variable \mbox{$(M_1,\dots,M_{n_m})$} whose elements are described by a categorical distribution \mbox{$m_i \sim \mathrm{Cat}(n_c, y_i)$, $i \in \{1\colon n_m\}$} where \mbox{$\mathrm{Cat}(n_c, y_i) = \prod_{k=1}^{n_c} y_i^{t_k}$}, $t_k \in \{0,1\}$,  and \mbox{$\sum_{k=1}^{n_c} t_k = 1$}. Given observations $Z= \boldsymbol z$, we wish to estimate \mbox{$p(M=m\mid Z= \boldsymbol z)$}.

Instead of solving the original multi-class classification problem directly, we further simplify the problem as follows. We use a binary RVM classifier, as described in the previous section, as the base classifier for each class and \emph{one-vs.-rest} approach to building a multi-class classifier. For any new input $\boldsymbol x_*$, the estimate of the posterior probability of class membership for class $c_k$, $k \in \{1:n_c\}$, can be computed as
\begin{equation}
\label{eq:rvsm_predict}
p( c_k = +1 | \Dcal, \boldsymbol x_*)  = \sigma(y(\boldsymbol x_* ; \boldsymbol w^{\star}))
\end{equation}

Once the $n_c$ binary RVM classifiers are trained, and the prediction at query points (map points) are performed, we normalize the class probabilities to get $p(M=c_k|\boldsymbol z)$ for the $k$-th semantic class. The straightforward way to assign hard labels (decisions) to map points is to find the class with the maximum probability. The actual representation of the map depends on the distribution of query points. In general, query points can have any desired distributions. However, in this work, we use the original dense point cloud as query points to facilitate comparison with ground truth.

\begin{algorithm}[t!]
\tiny{
\caption{\texttt{RVSM-training}}
\label{al:sequential}
\begin{algorithmic}[1]
\Require training set \mbox{$\Dcal = \{\boldsymbol x_i, t_i\}_{i=1}^{n_t}$}, semantic labels \mbox{$\mathcal{C} = \{\boldsymbol c_k\}_{k=1}^{n_c}$};
\State $\boldsymbol X \gets [\boldsymbol x_1, \dots, \boldsymbol x_{n_t}]$ // Matrix of inputs
\State // Train a binary RVM classifier for each semantic class
\For{$k \in \{1 : n_c\}$}
\State // Prepare input vector and target vector for class $k$
\For{$i \in \{1 : n_t\}$}
	\If{$t_i = c_k$}
    	\State $t_i^k = 1$
    \Else
    	\State $t_i^k = 0$
    \EndIf
\EndFor
\State $\boldsymbol t \gets \mathrm{vec}(t_1^k,\dots,t_{n_t}^k)$

\State // Initialize a single basis vector and its hyperparameter, and set all other hyperparameters to infinity
\State $\boldsymbol \phi_{1} \gets \boldsymbol \phi_{init}, \alpha_{1} \gets \frac{||\boldsymbol \phi_{1}||^{2}}{||\boldsymbol \phi_{1}^\transpose \boldsymbol t||^2 / ||\boldsymbol \phi_{1}||^{2} }$
\For{$m\in\{2:n_b\}$}
	\State $\alpha_{m} \gets \infty$
\EndFor

\State // Update model with initial basis and hyperparameter
\State $\mu, \Sigma, \boldsymbol s, \boldsymbol q \gets \texttt{Update-RVM-model}(\boldsymbol X, \boldsymbol t, \boldsymbol \phi_{1}, \alpha_{1})$

\State // Sequentially add and delete candidate basis vectors
\While{true}
\State $j \gets \texttt{Random}(1, n_b)$ // Randomly select a basis vector
\State $\theta_{j} \gets q_{j}^2 - s_{j}$
\If{$\theta_{j} > 0 $ and $\alpha_{j} < \infty$}
// $\boldsymbol \phi_{j}$ is in the current model\label{line:start_if}
	\State $\alpha_j \gets \texttt{Evaluate-hyperparameter}(s_j, q_j)$
\ElsIf{$\theta_{j} > 0 $ and $\alpha_{j} = \infty$}
// Add $\boldsymbol \phi_{j}$ to the model
	\State $\boldsymbol \phi_j \gets \boldsymbol \phi_{init}$
	\State $\alpha_j \gets \texttt{Evaluate-hyperparameter}(s_j, q_j)$
\ElsIf{$\theta_{j} \leq 0 $ and $\alpha_{j} < \infty$} // Delete $\boldsymbol \phi_j$ from the current model
	\State $\alpha_{j} \gets \infty$
\EndIf  \label{line:end_if}

\State // Update the current model at each iteration
\State $\boldsymbol \Phi \gets \varnothing, \boldsymbol \alpha \gets \varnothing$
\For{$m\in\{1:n_b\}$}
	\If{$\alpha_m < \infty$}
	\State \mbox{$\boldsymbol \Phi \gets \texttt{append}(\boldsymbol \Phi, \boldsymbol \phi_m), \boldsymbol \alpha \gets \texttt{append}(\boldsymbol \alpha, \alpha_m)$}
	\EndIf
\EndFor
\State $\boldsymbol \mu, \boldsymbol \Sigma, \boldsymbol s, \boldsymbol q \gets \texttt{Update-RVM-model}(\boldsymbol X, \boldsymbol t, \boldsymbol \Phi, \boldsymbol \alpha)$  \label{line:update}
\State // Output the model for class $k$
\If{\texttt{Converge}($\boldsymbol \alpha, \boldsymbol \theta$)}
	\State $\boldsymbol w^{\star}_k \gets \boldsymbol \mu, \boldsymbol \Sigma^{\star}_k \gets \boldsymbol \Sigma$
    \State \textbf{break}
\EndIf
\EndWhile
\EndFor
\State \Return $\mathcal{W} \gets \{\boldsymbol w^{\star}_k, \boldsymbol \Sigma^{\star}_k \}_{k=1}^{n_c}$ 
\end{algorithmic}}
\end{algorithm}

\begin{algorithm}[th!]
\tiny{
\caption{\texttt{Update-RVM-model}}
\label{al:updatemodel}
\begin{algorithmic}[1]
\Require Matrix of inputs $\boldsymbol X$, target vector $\boldsymbol t$, design matrix $\boldsymbol \Phi$, hyperparameters $\boldsymbol \alpha$;
\State // Find the Laplace approximation around the current mode
\State $\boldsymbol A \gets \diag{(\alpha_1, ..., \alpha_{n})}$ // $n$ is the number of hyperparameters in the current model
\For{$i\in\{1:n_t\}$}
	\State $\beta_i = \sigma \{ y(\boldsymbol x_i)\} \big[ 1 - \sigma \{ y( \boldsymbol x_i)\}]$
\EndFor
\State $\boldsymbol B \gets \diag{(\beta_1, ..., \beta_{n_t})}$
\State $\boldsymbol \mu = \boldsymbol A^{-1} \boldsymbol \Phi^\transpose (\boldsymbol t - \boldsymbol y)$ \quad // Equation~\eqref{eq:fast_mu}
\State $\boldsymbol \Sigma = (\boldsymbol \Phi^\transpose \boldsymbol B \boldsymbol \Phi + \boldsymbol A)^{-1}$ // Equation~\eqref{eq:fast_sigma}
\State $\hat{\boldsymbol t} \gets \boldsymbol \Phi \boldsymbol \mu + \boldsymbol B^{-1} (\boldsymbol t - \boldsymbol y)$
\State // Compute \emph{sparsity factor} and \emph{quality factor} based on the current approximation
\For{$m\in\{1:n_b\}$}
	\State $s \gets \boldsymbol \phi_m^{\transpose} \boldsymbol B \boldsymbol \phi_m - \boldsymbol \phi_m^{\transpose} \boldsymbol B \boldsymbol \Phi \boldsymbol \Sigma \boldsymbol \Phi^{\transpose} \boldsymbol B \boldsymbol \phi_m $
    \State $q \gets \boldsymbol \phi_m^{\transpose} \boldsymbol B \hat{\boldsymbol t} - \boldsymbol \phi_m^{\transpose} \boldsymbol B \boldsymbol \Phi \boldsymbol \Sigma \boldsymbol \Phi^{\transpose} \boldsymbol B \hat{\boldsymbol t}$
    \State // Sparsity and quality factors of basis $m$
    \State $s_m \gets \alpha_m s / (\alpha_m - s)$, $q_m \gets \alpha_m q / (\alpha_m - s)$ 
\EndFor
  \State \Return $\boldsymbol \mu, \boldsymbol \Sigma, \boldsymbol s \triangleq \mathrm{vec}(s_1,\dots,s_{n_b}), \boldsymbol q \triangleq \mathrm{vec}(q_1,\dots,q_{n_b})$
\end{algorithmic}}
\end{algorithm}

\begin{algorithm}[th!]
\tiny{
\caption{\texttt{Evaluate-hyperparameter}}
\label{al:evalalpha}
\begin{algorithmic}[1]
\Require sparsity factor $s$, quality factor $q$;
\If{$q^2 > s$} // The marginal likelihood maximized at finite $\alpha$
	\State $\alpha \gets s^2 / (q^2 - s)$
\Else \ // The marginal likelihood maximized as $\alpha \to \infty$
	\State $\alpha \gets \infty$
\EndIf
\State \Return $\alpha$
\end{algorithmic}}
\end{algorithm}

Algorithm \ref{al:sequential} shows the RVSM training algorithm. The algorithm exploits the sparsity of RVM in the training procedure by adding/deleting basis functions to maximize the marginal likelihood in~\eqref{eq:sbm_marginal} sequentially. In fact, this algorithm is guaranteed to maximize the marginal likelihood at each iteration~\cite{tipping2001sparse}. The algorithm avoids manipulating all basis functions at each iteration as in the original RVM algorithm, and thus accelerates the training process significantly. In this algorithm, a \emph{sparsity factor} and a \emph{quality factor} are defined as measures to evaluate each basis function, and the maximizing value of hyperparameter can be directly evaluated based on these two factors.  The update and hyperparameters evaluation methods are given in Algorithms~\ref{al:updatemodel} and~\ref{al:evalalpha} and explained as follows.

\texttt{Update-RVM-model}: Given the design matrix $\boldsymbol \Phi$ and the hyperparameter vector $\boldsymbol \alpha$, which contain only those basis vectors that are included in the current model, this function re-fits the Laplace approximation to the model, and computes the sparsity and quality factors for all $n_b$ bases. \texttt{Evaluate-hyperparameter}: This function estimates a hyperparameter based on its associated sparsity and quality factors. \texttt{Converge}: This function determines if the algorithm converges to a local maximum of the marginal likelihood by checking if the changes in $\boldsymbol \alpha$ are ``small enough''. When the algorithm converges, the marginal likelihood is maximized, and the hyperparameters and parameters in that iteration are $\boldsymbol \alpha^{\star}$ and $\boldsymbol w^{\star}$, respectively. Then, we can use the optimized values for prediction as shown in Algorithm~\ref{al:query}.

\section{Computational Complexity Analysis}
\label{sec:complexity}
The original RVM training time complexity is cubic in the number of basis functions, $n_b$, which initially starts with the number of training data, $n_t$, and rapidly reduces to less. This is because at convergence only a small number of basis functions survive (the relevance vectors). However, this is still an expensive process as $n_t$ can be large in dense robotic mapping problems. The sequential inference algorithm, Algorithm \ref{al:sequential}, speeds up the training process by initializing the model with zero bases and adding/deleting bases incrementally. The number of basis functions remains small during the entire inference, which reduces the time complexity significantly. The computational complexity of the RVSM for $n_c$ classes is therefore $\bigO{n_c n_i n_b^3}$ where $n_i$ is the number of iterations before convergence. Predictions in RVSM are linear in the number of relevance vectors, resulting in $\bigO{n_q n_c n_b}$ where $n_q$ is the number of queries.

\begin{algorithm}[t]
\tiny{
\caption{\texttt{RVSM-querying}}
\label{al:query}
\begin{algorithmic}[1]
\Require query points $\Xcal_* = \{\boldsymbol x_i\}_{i=1}^{n_q}$, training set $\Dcal$, trained model $\mathcal{W}$;
\For{$i \in \{1:n_q\}$}
    \State // Predictive probability of a query point taking label $c_k$
    \For{$k \in \{1:n_c\}$}
    	\State $p( c_k = +1 | \Dcal, \boldsymbol x_i)  \gets \sigma(y(\boldsymbol x_i; \boldsymbol w^{\star}_k))$ // Equation~\eqref{eq:rvsm_predict}
    \EndFor
    \For{$k \in \{1:n_c\}$} // Normalization
    	\State $p( m_i = c_k) \gets p( c_k = +1) / \sum_{k=1}^{n_c} p(c_k = +1)$
    \EndFor
\EndFor
\State \Return $p(M=m | Z = \boldsymbol z)$
\end{algorithmic}}
\end{algorithm}

\section{Results and Discussion}
\label{sec:results}
We evaluate the proposed semantic mapping algorithm using both indoor and outdoor datasets that are publicly available benchmark datasets, NYUDv2~\cite{Silberman:ECCV12} and KITTI~\cite{Geiger2012CVPR}. This section first explains the experimental setup and evaluation criteria used for the comparison, and then presents the qualitative and quantitative mapping results on both datasets. A discussion is also given at the end of this section.

\subsection{Experimental Setup and Evaluation Criteria}
\label{subsec:setup}
NYUDv2 provides a set of 1449 densely labeled \mbox{RGB-D} images that is split into a subset of 795 images for training/validation and 654 images for testing. The original dataset provides the image annotation for more than 1000 classes. 
We use the common 13-class task in our experiments to compare with available semantic mapping systems~\cite{hermans2014dense,wolf2015fast,mccormac2017semanticfusion}. The KITTI dataset does not offer semantic segmentation benchmark. In our experiments, we use the annotated KITTI odometry dataset provided in~\cite{sengupta2013urban}. For both datasets, we use deep semantic segmentation networks to compute pixel-wise labels for images and back-project the labels to their corresponding 3D points. The noisy labeled point clouds serve as observation sets to build semantic maps using RVSM.

To evaluate the mapping performance of RVSM and compare with the state-of-the-art semantic mapping systems, two metrics are used to quantify the results, \mbox{$\text{Sensitivity (Recall)} = \frac{\mathrm{TP}}{\mathrm{TP}+\mathrm{FN}}$} and the \emph{Area Under the receiver operating characteristic Curve} (AUC), where $\transpose/\mathrm{F}$ and $\mathrm{P}/\mathrm{N}$ denote true/false and positive/negative, respectively. As the RVSM gives probabilistic outputs, the AUC is the appropriate metric to measure the mapping performance. The AUC handles probabilistic outputs (soft labels) of a classifier and does not resort to a fixed threshold to make decisions. Furthermore, the AUC maintains the performance measure insensitive to skewed class distributions and error costs~\cite{fawcett2006introduction}. For a fair comparison, we also compute the Sensitivity as reported in other works. We calculate these metrics for each semantic class and report the class average results. As the Sensitivity cannot deal with probabilistic outputs, we report the mean of this metric over the range of a decision threshold interval $(0,1)$.

\begin{figure*}[t!]
  \centering
    \subfloat{\includegraphics[width=.5\columnwidth]{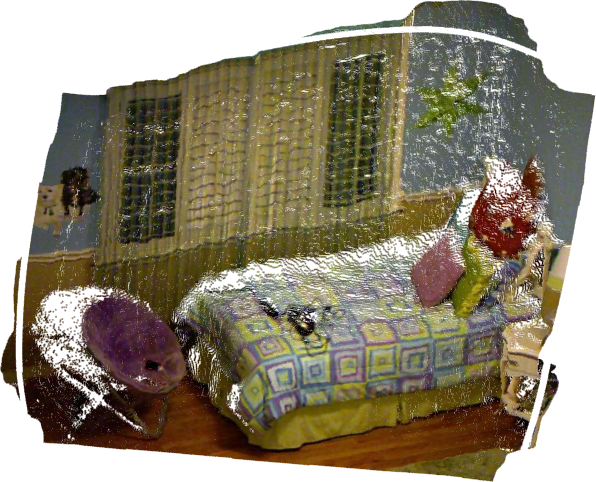}
  \label{fig:gt2}}
  \subfloat{\includegraphics[width=.5\columnwidth]{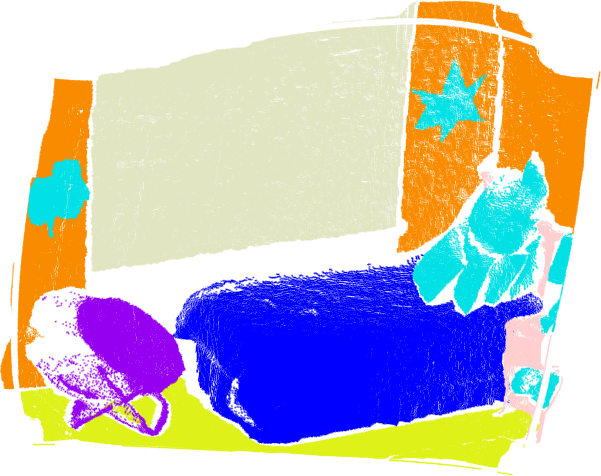}
  \label{fig:gtd2}}
  \subfloat{\includegraphics[width=.5\columnwidth]{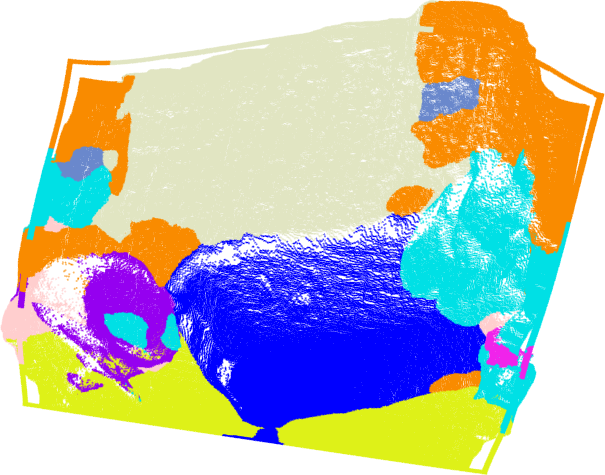}
  \label{fig:som2}}
  \subfloat{\includegraphics[width=.5\columnwidth]{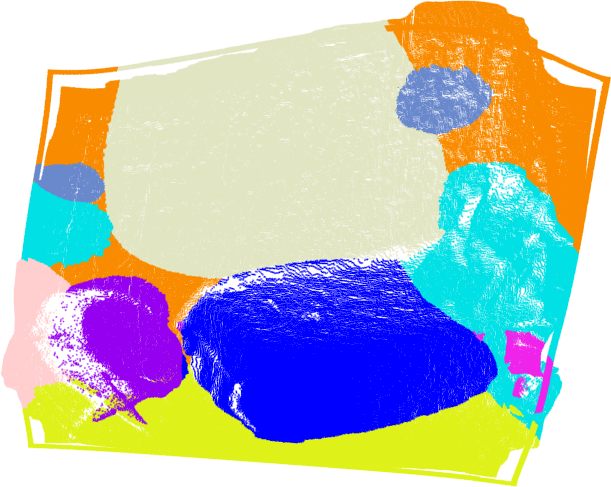}
  \label{fig:gpsm2}} \\
  
    \subfloat{\includegraphics[width=0.5\columnwidth]{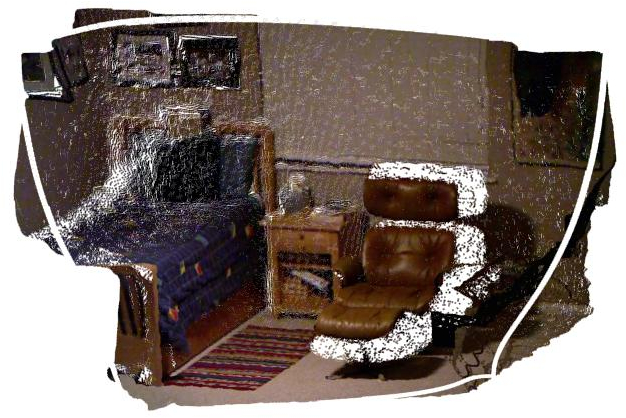}
    \label{fig:gt_c2}}
  \subfloat{\includegraphics[width=0.5\columnwidth]{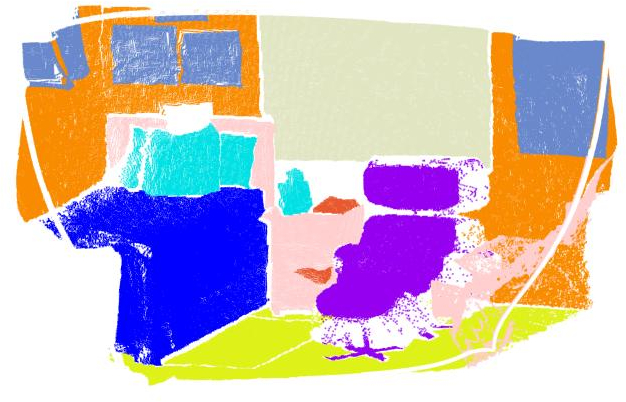}
    \label{fig:gtd_c2}}
  \subfloat{\includegraphics[width=0.5\columnwidth]{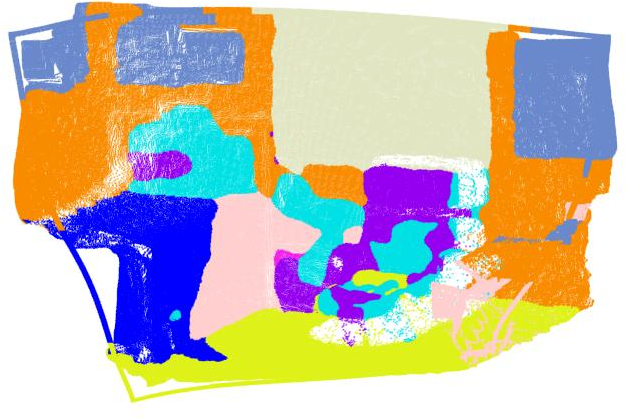}
    \label{fig:som_c2}}
  \subfloat{\includegraphics[width=0.5\columnwidth]{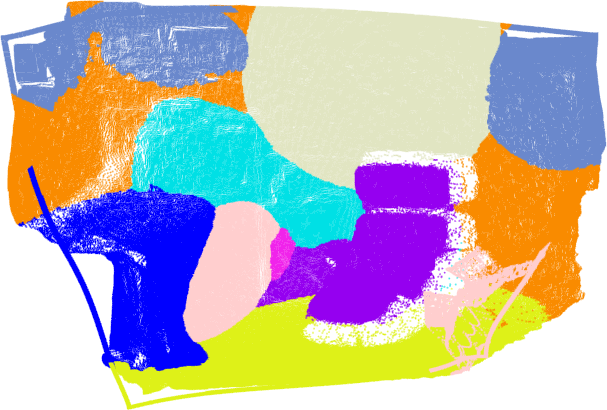}
    \label{fig:gpsm_c2}} \\

\crule[bedColor]{0.4cm}{0.3cm} \scriptsize{Bed}
\crule[booksColor]{0.4cm}{0.3cm} \scriptsize{Books}
\crule[ceilColor]{0.4cm}{0.3cm} \scriptsize{Ceiling}
\crule[chairColor]{0.4cm}{0.3cm} \scriptsize{Chair}
\crule[floorColor]{0.4cm}{0.3cm} \scriptsize{Floor}
\crule[furnColor]{0.4cm}{0.3cm} \scriptsize{Furniture}
\crule[objsColor]{0.4cm}{0.3cm} \scriptsize{Objects}
\crule[paintColor]{0.4cm}{0.3cm} \scriptsize{Painting}
\crule[sofaColor]{0.4cm}{0.3cm} \scriptsize{Sofa}
\crule[tableColor]{0.4cm}{0.3cm} \scriptsize{Table}
\crule[tvColor]{0.4cm}{0.3cm} \scriptsize{TV}
\crule[wallColor]{0.4cm}{0.3cm} \scriptsize{Wall}
\crule[windColor]{0.4cm}{0.3cm} \scriptsize{Window}
  \caption{Qualitative results of mapping under noisy and misclassified labels on NYUDv2 test set~\cite{Silberman:ECCV12} for 13 semantic classes. From left, each column respectively shows the input point clouds, the ground truth labels, the segmentation results by Eigen \textit{et al.}~\cite{eigen2015predicting}, and the RVSM results.}
  \label{fig:nyu_rvsm}
  \squeezeup
\end{figure*}

\begin{table*}[t]
\centering
\caption{Quantitative results on NYUDv2 test set~\cite{Silberman:ECCV12} for 13 semantic classes. Sensitivity and AUC are evaluated for the comparison of 2D/3D methods.}
\resizebox{\textwidth}{!}{
\begin{tabular}{lllcccccccccccccc}
Metric & \multicolumn{2}{c|}{Method}  & \cellcolor{bedColor}\rotatebox{90}{\color{white}Bed} & \cellcolor{booksColor}\rotatebox{90}{Books}  & \cellcolor{ceilColor}\rotatebox{90}{Ceiling} & \cellcolor{chairColor}\rotatebox{90}{\color{white}Chair}  & \cellcolor{floorColor}\rotatebox{90}{Floor}  & \cellcolor{furnColor}\rotatebox{90}{Furn.}   & \cellcolor{objsColor}\rotatebox{90}{Objects} & \cellcolor{paintColor}\rotatebox{90}{Painting} & \cellcolor{sofaColor}\rotatebox{90}{\color{white}Sofa}   & \cellcolor{tableColor}\rotatebox{90}{Table}  & \cellcolor{tvColor}\rotatebox{90}{TV}     & \cellcolor{wallColor}\rotatebox{90}{Wall}   & \cellcolor{windColor}\rotatebox{90}{Window} &
\rotatebox{90}{Average} \\ \hline

\vspace{-2mm} \\

\multirow{5}{*}{Sensitivity (Recall)} & \multirow{1}{*}{2D} & Eigen \textit{et al.} \cite{eigen2015predicting} 
& 66.5 & 67.0 & 80.1 & 73.1 & 95.5 & 58.8 & 42.7 & 69.3 & 59.3 & 53.5 & 61.4 & 90.0 & 69.5 & 68.2 \\ \cline{2-17} 
& \multirow{4}{*}{3D}
& SemanticFusion \cite{mccormac2017semanticfusion} & 48.3 & 51.5 & 79.0 & \textbf{74.7} & 90.8 & \textbf{63.5} & \textbf{46.9} & 63.6 & 46.5 & 45.9 & \textbf{71.5} & \textbf{89.4} & 55.6 & 63.6 \\
& & MVCNet-MaxPool \cite{ma2017multi} & 65.7 & 49.2 & 66.4 & 54.6 & 89.9 & 59.2 & 39.1 & 49.5 & 56.3 & 43.5 & 37.4 & 75.3 & 59.1 & 57.3 \\
& & Wolf \textit{et al.} \cite{wolf2015fast} & 58.2 & 45.3 & \textbf{92.8} & 54.7 & \textbf{97.5} & 57.3 & 37.4 & 32.3 & 49.8 & 51.8 & 26.4 & 74.4 & 43.2 & 55.5\\
& & RVSM & \textbf{67.4}  & \textbf{71.2}  & 83.6 & 67.8 & 92.7 & 54.4 & 38.6 & \textbf{71.0} & \textbf{61.6} & \textbf{53.6}  & 69.2 & 77.7 & \textbf{73.6} & \textbf{67.9} \\ \midrule

\multirow{2}{*}{AUC} & \multirow{1}{*}{2D} & Eigen \textit{et al.} \cite{eigen2015predicting} 
& 81.1 & 78.3 & 88.8 & 84.0 & 96.6 & 74.2 & 67.8 & 83.2 & 77.7 & 75.4 & 80.1 & 86.3 & 83.2 & 81.3\\ \cline{2-17} 
& 3D &RVSM
& \textbf{88.4} & \textbf{86.9} & \textbf{94.4} & \textbf{90.2} & \textbf{98.8} & \textbf{80.9} & \textbf{73.4} & \textbf{89.6} & \textbf{86.7} & \textbf{85.3} & \textbf{90.3} & \textbf{92.2} & \textbf{91.0} & \textbf{88.4} \\ \bottomrule

\end{tabular}}
\label{table:NYU13classes}
\squeezeup
\end{table*}

\begin{table*}[t]
\centering
\caption{Comparison of the average number of points in the original point cloud, points in the training set, and relevance vectors used for prediction per semantic class for NYUDv2 test set~\cite{Silberman:ECCV12}.}
\resizebox{\textwidth}{!}{
\begin{tabular}{lrrrrrrrrrrrrrrrr}
Average Number of & \cellcolor{bedColor}\rotatebox{90}{\color{white}Bed} & \cellcolor{booksColor}\rotatebox{90}{Books}  & \cellcolor{ceilColor}\rotatebox{90}{Ceiling} & \cellcolor{chairColor}\rotatebox{90}{\color{white}Chair}  & \cellcolor{floorColor}\rotatebox{90}{Floor}  & \cellcolor{furnColor}\rotatebox{90}{Furn.}   & \cellcolor{objsColor}\rotatebox{90}{Objects} & \cellcolor{paintColor}\rotatebox{90}{Painting} & \cellcolor{sofaColor}\rotatebox{90}{\color{white}Sofa}   & \cellcolor{tableColor}\rotatebox{90}{Table}  & \cellcolor{tvColor}\rotatebox{90}{TV}     & \cellcolor{wallColor}\rotatebox{90}{Wall}   & \cellcolor{windColor}\rotatebox{90}{Window} &
\rotatebox{90}{Average} \\ \hline
\vspace{-2mm} \\
\multirow{1}{*}{Measurements (Points)} & 35648 & 18988 & 16703 & 21184 & 40662  &  48163 &  33497 & 13567  & 30174 & 16509  &   5716 & 100298  & 25044  &  31243 \\ 
\multirow{1}{*}{Positive Training Instances} & 102 & 63 & 84 & 97  & 133 & 120 & 133 & 89 & 100 & 90 & 43 & 149 & 104 & 101  \\ 
\multirow{1}{*}{Negative Training Instances} & 862 & 943 & 909 & 899  & 825 & 813 & 795 & 867 & 941 & 925 & 915 & 778 & 872 & 873 \\ 
\multirow{1}{*}{Relevance Vectors} & 5 & 4 & 4 & 7 & 4 & 7 & 8 & 5 & 6 & 6 & 4 & 9 & 5 & 6 \\ \bottomrule

\end{tabular}}
\label{table:NYU_points}
\squeezeup\squeezeup
\end{table*}

\subsection{NYU Dataset}
\label{subsec:nyuresult}

The NYUDv2 dataset comprises video sequences recorded by Microsoft Kinect camera with $640 \times 480$ resolution from a variety of indoor scenes such as bedroom, office, kitchen and living room. In this experiment, we build the RVSM for each frame using the 654 test images, and evaluate the mapping performance on the entire test set using ground truth labels.

We first generate point clouds from inpainted depth images using the MATLAB toolbox provided with the dataset, and label the point clouds by transferring 2D pixel-wise labels. The pixel-wise labels are obtained from the image semantic labeling results of a multi-scale deep convolutional network proposed by Eigen \textit{et al.}~\cite{eigen2015predicting}. This work achieves excellent labeling performance on NYUDv2 13-class task, and provides the trained models along with the predicted outputs. We directly get the predicted semantic labels for 13 classes using their VGG-based model, which has better accuracy than their AlexNet-based model. The model is trained on the training set and the predicted semantic labels are given for all 654 test images. In view of fast inference, we uniformly downsample the labeled point clouds by keeping one percent of the data for each class. Then, we use the downsampled point clouds with noisy labels to build the RVSM. 

Fig.~\ref{fig:nyu_rvsm} shows the examples of results using NYUDv2, where the RVSM result is shown along with the corresponding input point cloud, ground truth labels, and point cloud labeled purely by Eigen's segmentation results. The RVSM results are visualized using the class labels with the maximum probability. As we can see, the RVSM can correct some of the misclassified labels in the map based on spatial correlations between map points. We compare the RVSM with approaches in~\cite{mccormac2017semanticfusion,ma2017multi,wolf2015fast}, and the quantitative results are given in Table \ref{table:NYU13classes}. 

SemanticFusion~\cite{mccormac2017semanticfusion} combines a deep segmentation network and a SLAM system to build a semantic map. Different segmentation networks are used in~\cite{mccormac2017semanticfusion}, and we report the best results obtained by the Eigen-SF-CRF model. However, their results are only evaluated on a subset of the test set and thus are not directly comparable. MVCNet~\cite{ma2017multi} is an end-to-end deep learning semantic mapping approach that fuses the semantic predictions of individual views into a common reference view to achieve multi-view consistency. For a fair comparison, we compare the RVSM with the best single-view method MVCNet-MaxPool. Our results are not directly comparable with Wolf \textit{et al.}~\cite{wolf2015fast} as \emph{bookshelf} and \emph{wall decoration} instead of \emph{books} and \emph{painting} are used in the annotations. However, we include their results for the reference. We note that the ``accuracy'' used in the referred results corresponds to the definition of Sensitivity (Recall) as reported here.

From Table~\ref{table:NYU13classes}, the RVSM has higher sensitivity in 6 out of 13 classes and on average. For the class \emph{books} and \emph{window}, there is a sensitivity increase of more than $10\%$ over other systems; however, we should acknowledge that a reason for this improvement can be the high sensitivity of the used 2D segmentation method. Furthermore, the RVSM improves the classification performance over 2D segmentation by a higher AUC compared with Eigen \textit{et al.}~\cite{eigen2015predicting}. RVSM infers class labels based on the available correlation in observation sets, and the spatial correlations in this case reduce the number of misclassified points in the final map; an advantage of high-dimensional inference. Table~\ref{table:NYU_points} summarizes the number of points in measurements, training set, and inferred relevance vectors, which indicates the sparsity of the RVSM.

\subsection{KITTI Dataset}
\label{subsec:hellokitti}

The KITTI odometry dataset consists of 22 outdoor-scene stereo sequences with $1226 \times 370$ resolution recorded by the sensor mounted on a driving vehicle. In this experiment, we build local RVSM for 25 test images in sequence 15, and quantitatively compare with other outdoor semantic mapping systems. More preprocessing work is required to build the RVSM for KITTI dataset. We first use LIBELAS \cite{Geiger2010ACCV} to get dense and accurate disparity maps using the rectified stereo images. Then, we back-project the disparity map to 3D space using camera matrices given in the dataset. For segmentation results, a deep Dilated CNN~\cite{YuKoltun2016} with trained model on KITTI dataset is adopted. The model is trained previously using 100 images in sequence 00, and is used to predict semantic labels for the 25 test images in sequence 15 similar to~\cite{yang2017semantic}. To compare with other methods, we build the RVM for 7 common semantic classes, i.e., \emph{building, vegetation, car, road, fence, sidewalk, pole}.

\begin{table}[t]
\centering
\caption{Quantitative results on KITTI odometry sequence 15 test set~\cite{sengupta2013urban} for 7 semantic classes. Sensitivity and the AUC are evaluated for the comparison of 2D/3D methods.}
\label{table:kitti15}
\resizebox{\columnwidth}{!}{
\begin{tabular}{lllcccccccc}
Metric & \multicolumn{2}{c}{Method}\textbf{}
& \cellcolor{buildingColor}\rotatebox{90}{\color{white}Building} & \cellcolor{vegetationColor}\rotatebox{90}{\color{white}Vege.} & \cellcolor{carColor}\rotatebox{90}{\color{white}Car} & \cellcolor{roadColor}\rotatebox{90}{\color{white}Road} & \cellcolor{fenceColor}\rotatebox{90}{\color{white}Fence} & \cellcolor{sidewalkColor}\rotatebox{90}{\color{white}Sidewalk} & \cellcolor{poleColor}\rotatebox{90}{\color{white}Pole} & \rotatebox{90}{Average} \\ \hline 

\vspace{-2mm} \\

\multirow{6}{*}{Sensitivity} & \multirow{2}{*}{2D} & Ladický \textit{et al.} \cite{ladicky2014associative}
& 97.0 & 93.4 & 93.9 & 98.3 & 48.5 & \textbf{91.3} & \textbf{49.3} & 81.7 \\
& & Dilated CNN \cite{YuKoltun2016} & \textbf{97.7} & \textbf{97.1} & \textbf{96.5} & \textbf{98.4} & \textbf{77.8} & 87.9 & 49.2 & \textbf{86.4} \\ \cline{2-11} 
\multirow{4}{*}{(Recall)}
& \multirow{5}{*}{3D}
& 	Valentin \textit{et al.} \cite{valentin2013mesh}& 96.4 & 85.4 	& 76.8 	& 96.9 	& 42.7 	& 78.5 	& 39.3 & 73.7 \\
& & Sengupta \textit{et al.} \cite{sengupta2013urban} 	& 96.1	& 86.9	& 88.5	& 97.8 & 46.1 	& 86.5 	& 38.2 	& 77.2 \\
& & Vineet \textit{et al.} \cite{vineet2015incremental} & 97.2	& 94.1 & 94.1 	& \textbf{98.7} & 47.8 	& 91.8	& 51.4 	& 82.2 \\
& & Semantic Octree \cite{sengupta2015semantic} 	& 89.1 	& 81.2 	& 72.5 	& 97.0 & 45.7 	& 73.4 	& 3.3 	& 66.0 \\
& & Yang \textit{et al.} \cite{yang2017semantic}	& \textbf{98.2} & \textbf{98.7} & \textbf{95.5} & \textbf{98.7} & 84.7 & \textbf{93.8} & 66.3 & \textbf{90.9} \\
& & RVSM & 97.8	& 90.9 & 89.2  & 96.5 	& \textbf{93.4}	& 90.1 & \textbf{69.7}	& 89.7 \\ \midrule

\multirow{2}{*}{AUC} & 2D & Dilated CNN \cite{YuKoltun2016} 
& 96.6 & 95.7 & \textbf{98.0} & 98.3 & 88.6 & 93.3 & 74.5 & 92.2 \\ \cline{2-11} 
& 3D & RVSM & \textbf{97.0} & 95.7 & 95.0 & \textbf{98.5} & \textbf{97.1} & \textbf{95.9} & \textbf{91.1} & \textbf{95.7} \\ \bottomrule
\end{tabular}}
\squeezeup\squeezeup
\end{table}

\begin{table}[t]
\centering
\caption{Comparison of the average number of points in the original point cloud, points in the training set, and relevance vectors used for prediction per semantic class for KITTI odometry sequence 15 test set~\cite{sengupta2013urban}.}
\label{table:kitti_points}
\resizebox{\columnwidth}{!}{
\begin{tabular}{lrrrrrrrrrr}
\multicolumn{1}{l|}{Average Number of}
& \cellcolor{buildingColor}\rotatebox{90}{\color{white}Building} & \cellcolor{vegetationColor}\rotatebox{90}{\color{white}Vege.} & \cellcolor{carColor}\rotatebox{90}{\color{white}Car} & \cellcolor{roadColor}\rotatebox{90}{\color{white}Road} & \cellcolor{fenceColor}\rotatebox{90}{\color{white}Fence} & \cellcolor{sidewalkColor}\rotatebox{90}{\color{white}Sidewalk} & \cellcolor{poleColor}\rotatebox{90}{\color{white}Pole} & \rotatebox{90}{Average} \\ \hline 
\vspace{-2mm} \\
\multirow{1}{*}{Measurements (Points)} & 70334 & 75804 & 25262 & 46168 & 29593 & 21493 & 1724 & 38625
\\ 
\multirow{1}{*}{Positive Training Instances} & 467 & 492 & 162 & 303 & 190 & 138 & 6 & 251
\\ 
\multirow{1}{*}{Negative Training Instances} & 1292 & 1267 & 1595 & 1455 & 1568 & 1621 & 1752 & 1507
\\ 
\multirow{1}{*}{Relevance Vectors} & 24 & 30 & 18 & 20 & 21 & 19 & 16 & 24
\\ \bottomrule
\end{tabular}}
\squeezeup\squeezeup
\end{table}

\begin{figure}[t]
\centering
\subfloat{\includegraphics[width=.96\columnwidth,trim={-1.05cm 0cm 0cm 0cm},clip]{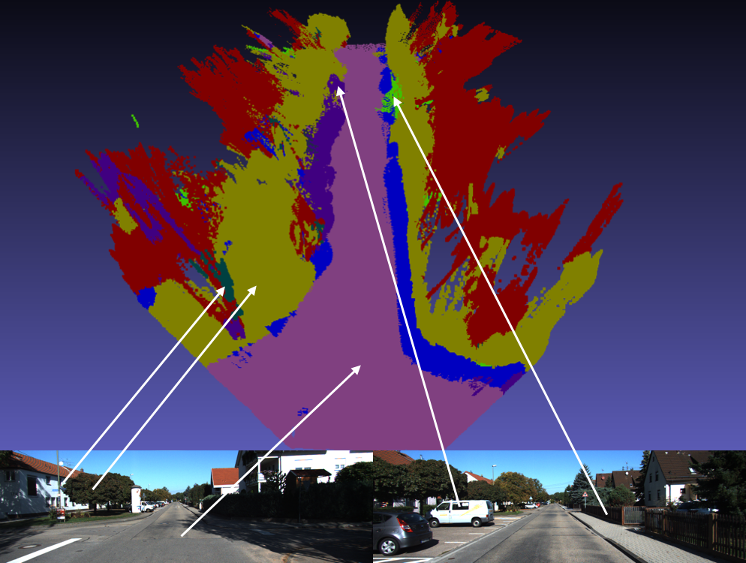}} \\
\squeezeup
\subfloat{\includegraphics[width=.95\columnwidth,trim={0cm -0.5cm 0cm 0cm},clip]{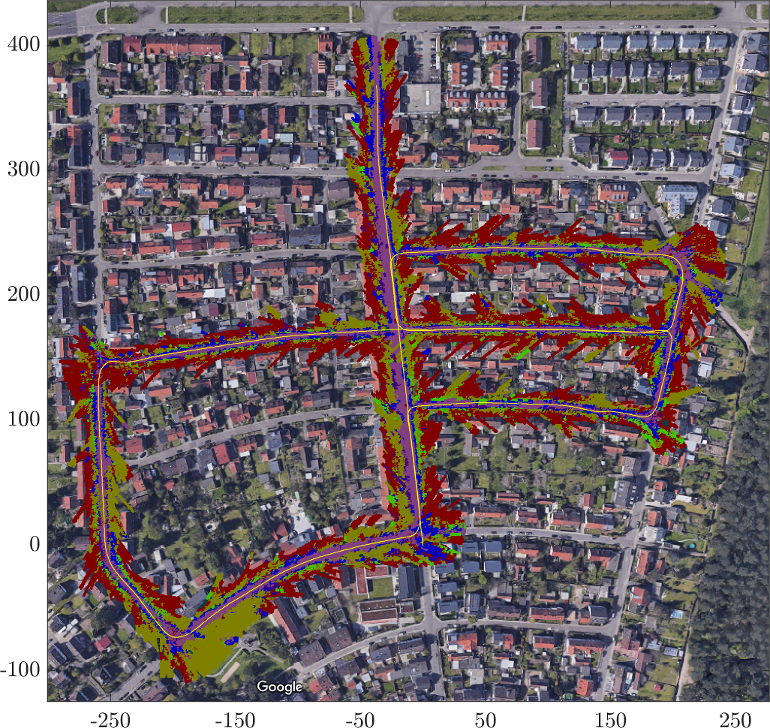}}\\
\crule[buildingColor]{0.4cm}{0.3cm} \scriptsize{Building}
\crule[vegetationColor]{0.4cm}{0.3cm} \scriptsize{Vegetation}
\crule[carColor]{0.4cm}{0.3cm} \scriptsize{Car}
\crule[roadColor]{0.4cm}{0.3cm} \scriptsize{Road}
\crule[fenceColor]{0.4cm}{0.3cm} \scriptsize{Fence}
\crule[sidewalkColor]{0.4cm}{0.3cm} \scriptsize{Sidewalk}
\crule[poleColor]{0.4cm}{0.3cm} \scriptsize{Pole}
\caption{Large-scale 3D dense RVSM over long sequence (a subset of 2760 images) of KITTI odometry sequence 05~\cite{Geiger2012CVPR}. The top figure shows a closeup view of the RVSM with corresponding images, where the white arrows show the correspondent objects between 3D semantic map and 2D images. The bottom figure shows a top view of the RVSM overlaid on Google satellite map, where the yellow line shows the ground truth trajectory in meters.}
\label{fig:kitti05}
\squeezeup\squeezeup
\end{figure}

Fig.~\ref{fig:first} shows an example of the qualitative RVSM results for sequence 15, along with the corresponding left image in the stereo pair and the segmentation result of Dilated CNN~\cite{YuKoltun2016}. The RVSM result is built using 10 successive frames in the test set by stitching local maps together using GICP algorithm~\cite{segal2009generalized}. Visually, we can see that the RVSM can capture details of the scene in spite of the small volume and texture-less surfaces.

Quantitative results are given in Table \ref{table:kitti15}. We compare the RVSM with other outdoor semantic mapping systems that report results on the same dataset. We project the RVSM onto the camera image plane, while discarding points that are further than 40 meters away from the camera, similar to~\cite{yang2017semantic}. In~\cite{vineet2015incremental} a 25 meters range is used to produce the results. The 2D method in~\cite{ladicky2014associative} is the segmentation algorithm used in 3D methods~\cite{sengupta2013urban,vineet2015incremental}, and Yang \textit{et al.} use Dilated CNN~\cite{YuKoltun2016} to get pixel-wise labels. Description of these 3D methods can be found in our related work, Section~\ref{sec:related}. From Table \ref{table:kitti15}, the RVSM has comparable sensitivity while better performance than 2D segmentation results. Table~\ref{table:kitti_points} summarizes the number of points in measurements, training set, and inferred relevance vectors.

A large-scale RVSM is also built using a subset of 2760 images in KITTI odometry sequence 05. We use the ground truth poses provided in the dataset to build the global semantic map for the entire sequence. The RVSM is shown in Fig.~\ref{fig:kitti05} with a closeup view and a top view overlaid on the Google satellite map of the area. The correspondences of semantic objects between the RVSM and 2D images are also highlighted. From Fig.~\ref{fig:kitti05}, we can see the effectiveness of the RVSM in reconstructing semantic objects in a large-scale outdoor scene over long sequence data.

\subsection{Discussion and Limitations}
\label{subsec:discussion}
Intuitively, it is easier to discriminate between points in a higher-dimensional space; therefore, given prior 2D labels, a better performance in 3D is expected. However, the presence of false positives in the image segmentation results degrades 3D mapping performance. A special case is where the segmented image contains false positive labels for classes that do not exist in the image. In such scenarios, training a classifier using those labels can considerably reduce the performance. A possible solution is exploiting the available time correlations between successive images to reject such outliers prior to inference. Fusing local RVSMs into a global map can also improve the performance.

In the current RVSM implementation, we use the kernel functions defined in our previous GP semantic map representation~\cite{gpsm2017rssws}. However, applying Bayesian model selection and jointly optimizing the kernel parameters in RVSM inference can improve the performance and are interesting future research directions.

\section{Conclusion and Future Work}
\label{sec:conclusion}
In this paper, we proposed a novel dense 3D semantic mapping algorithm using a sparse Bayesian model, the relevance vector machine. We formulated the problem as a high-dimensional multi-class classification and solved it sequentially in a fully probabilistic framework. The proposed RVSM is continuous and predictions can be made efficiently at any desired point in the map. The provided evaluations in indoor and outdoor benchmark datasets showed that RVSM brings many desirable features of fully Bayesian frameworks such as GPs while it is sparse and its performance is comparable with the current available systems for dense semantic mapping.

While we simplified the multi-class problem into a multiple binary classifications in this paper, we acknowledge that the likelihood can be generalized to a multinomial model to integrate an inherently multi-class classifier into the RVSM framework. In addition, our future work includes exploring additional non-spatial correlation, establishing a probabilistic map fusion framework, and implementing RVSM on a mobile robotic platform for real-time experiments. 



\bibliographystyle{IEEEtran} 
\bibliography{references}

\begin{thebibliography}{10}
\providecommand{\url}[1]{#1}
\csname url@rmstyle\endcsname
\providecommand{\newblock}{\relax}
\providecommand{\bibinfo}[2]{#2}
\providecommand\BIBentrySTDinterwordspacing{\spaceskip=0pt\relax}
\providecommand\BIBentryALTinterwordstretchfactor{4}
\providecommand\BIBentryALTinterwordspacing{\spaceskip=\fontdimen2\font plus
\BIBentryALTinterwordstretchfactor\fontdimen3\font minus
  \fontdimen4\font\relax}
\providecommand\BIBforeignlanguage[2]{{%
\expandafter\ifx\csname l@#1\endcsname\relax
\typeout{** WARNING: IEEEtran.bst: No hyphenation pattern has been}%
\typeout{** loaded for the language `#1'. Using the pattern for}%
\typeout{** the default language instead.}%
\else
\language=\csname l@#1\endcsname
\fi
#2}}

\bibitem{Thrun:2003:RMS:779343.779345}
S.~Thrun, ``Exploring artificial intelligence in the new millennium,''
  G.~Lakemeyer and B.~Nebel, Eds.\hskip 1em plus 0.5em minus 0.4em\relax Morgan
  Kaufmann Publishers Inc., 2003, ch. Robotic Mapping: A Survey, pp. 1--35.

\bibitem{hornung2013octomap}
A.~Hornung, K.~M. Wurm, M.~Bennewitz, C.~Stachniss, and W.~Burgard,
  ``{OctoMap}: An efficient probabilistic {3D} mapping framework based on
  octrees,'' \emph{Auton. Robot}, vol.~34, no.~3, pp. 189--206, 2013.

\bibitem{merali2014optimizing}
R.~S. Merali and T.~D. Barfoot, ``Optimizing online occupancy grid mapping to
  capture the residual uncertainty,'' in \emph{Proc. IEEE Int. Conf. Robot
  Automat.}, 2014, pp. 6070--6076.

\bibitem{charrow2015information}
B.~Charrow, S.~Liu, V.~Kumar, and N.~Michael, ``Information-theoretic mapping
  using {Cauchy-Schwarz} quadratic mutual information,'' in \emph{Proc. IEEE
  Int. Conf. Robot Automat.}, 2015, pp. 4791--4798.

\bibitem{bowman2017probabilistic}
S.~L. Bowman, N.~Atanasov, K.~Daniilidis, and G.~J. Pappas, ``Probabilistic
  data association for semantic {SLAM},'' in \emph{Proc. IEEE Int. Conf. Robot
  Automat.}, 2017, pp. 1722--1729.

\bibitem{runz2017co}
M.~R{\"u}nz and L.~Agapito, ``Co-fusion: Real-time segmentation, tracking and
  fusion of multiple objects,'' in \emph{Proc. IEEE Int. Conf. Robot Automat.},
  2017, pp. 4471--4478.

\bibitem{eigen2015predicting}
D.~Eigen and R.~Fergus, ``Predicting depth, surface normals and semantic labels
  with a common multi-scale convolutional architecture,'' in \emph{Proc. IEEE
  Int. Conf. Computer Vision}, 2015, pp. 2650--2658.

\bibitem{YuKoltun2016}
F.~Yu and V.~Koltun, ``Multi-scale context aggregation by dilated
  convolutions,'' in \emph{Proc. Int. Conf. Learn. Represent.}, Vancouver, BC,
  Canada, 2016, pp. 1--13.

\bibitem{valentin2013mesh}
J.~P. Valentin, S.~Sengupta, J.~Warrell, A.~Shahrokni, and P.~H. Torr, ``Mesh
  based semantic modelling for indoor and outdoor scenes,'' in \emph{Proc. IEEE
  Int. Conf. Computer Vision and Pattern Recog.}, Portland, Oregon, 2013, pp.
  2067--2074.

\bibitem{sengupta2013urban}
S.~Sengupta, E.~Greveson, A.~Shahrokni, and P.~H. Torr, ``Urban {3D} semantic
  modelling using stereo vision,'' in \emph{Proc. IEEE Int. Conf. Robot
  Automat.}, 2013, pp. 580--585.

\bibitem{ladicky2014associative}
L.~Ladick{\`y}, C.~Russell, P.~Kohli, and P.~H. Torr, ``Associative
  hierarchical random fields,'' \emph{IEEE Trans. Pattern Anal. Machine
  Intell.}, vol.~36, no.~6, pp. 1056--1077, 2014.

\bibitem{vineet2015incremental}
V.~Vineet, O.~Miksik, M.~Lidegaard, M.~Nie{\ss}ner, S.~Golodetz, V.~A.
  Prisacariu, O.~K{\"a}hler, D.~W. Murray, S.~Izadi, P.~P{\'e}rez,
  \emph{et~al.}, ``Incremental dense semantic stereo fusion for large-scale
  semantic scene reconstruction,'' in \emph{Proc. IEEE Int. Conf. Robot
  Automat.}, 2015, pp. 75--82.

\bibitem{sengupta2015semantic}
S.~Sengupta and P.~Sturgess, ``Semantic octree: Unifying recognition,
  reconstruction and representation via an octree constrained higher order
  {MRF},'' in \emph{Proc. IEEE Int. Conf. Robot Automat.}, 2015, pp.
  1874--1879.

\bibitem{wolf2015fast}
D.~Wolf, J.~Prankl, and M.~Vincze, ``Fast semantic segmentation of {3D} point
  clouds using a dense {CRF} with learned parameters,'' in \emph{Proc. IEEE
  Int. Conf. Robot Automat.}, 2015, pp. 4867--4873.

\bibitem{mccormac2017semanticfusion}
J.~McCormac, A.~Handa, A.~Davison, and S.~Leutenegger, ``Semanticfusion: Dense
  {3D} semantic mapping with convolutional neural networks,'' in \emph{Proc.
  IEEE Int. Conf. Robot Automat.}, 2017, pp. 4628--4635.

\bibitem{ma2017multi}
L.~Ma, J.~St{\"u}ckler, C.~Kerl, and D.~Cremers, ``Multi-view deep learning for
  consistent semantic mapping with {RGB-D} cameras,'' \emph{arXiv:1703.08866},
  2017.

\bibitem{yang2017semantic}
S.~Yang, Y.~Huang, and S.~Scherer, ``Semantic {3D} occupancy mapping through
  efficient high order {CRF}s,'' \emph{arXiv:1707.07388}, 2017.

\bibitem{murphy2012machine}
K.~P. Murphy, \emph{Machine learning: a probabilistic perspective}.\hskip 1em
  plus 0.5em minus 0.4em\relax The MIT Press, 2012.

\bibitem{paul2012semantic}
R.~Paul, R.~Triebel, D.~Rus, and P.~Newman, ``Semantic categorization of
  outdoor scenes with uncertainty estimates using multi-class {Gaussian}
  process classification,'' in \emph{Proc. IEEE/RSJ Int. Conf. Intell. Robots
  Syst.}, 2012, pp. 2404--2410.

\bibitem{gpsm2017rssws}
M.~Ghaffari~Jadidi, L.~Gan, S.~A. Parkison, J.~Li, and R.~M. Eustice,
  ``{Gaussian} processes semantic map representation,'' in \emph{{RSS} Workshop
  on Spatial-Semantic Representations in Robotics}, 2017.

\bibitem{Geiger2012CVPR}
A.~Geiger, P.~Lenz, and R.~Urtasun, ``Are we ready for autonomous driving? the
  {KITTI} vision benchmark suite,'' in \emph{Proc. IEEE Int. Conf. Computer
  Vision and Pattern Recog.}, 2012.

\bibitem{tipping2001sparse}
M.~E. Tipping, ``Sparse {Bayesian} learning and the relevance vector machine,''
  \emph{J. machine learning res.}, vol.~1, no. Jun, pp. 211--244, 2001.

\bibitem{tipping2003fast}
M.~E. Tipping, A.~C. Faul, \emph{et~al.}, ``Fast marginal likelihood
  maximisation for sparse bayesian models.'' in \emph{AISTATS}, 2003.

\bibitem{neal1996bayesian}
R.~M. Neal, \emph{{Bayesian} learning for neural networks}.\hskip 1em plus
  0.5em minus 0.4em\relax Springer New York, 1996, vol. 118.

\bibitem{he2013nonparametric}
H.~He and B.~Upcroft, ``Nonparametric semantic segmentation for {3D} street
  scenes,'' in \emph{Proc. IEEE/RSJ Int. Conf. Intell. Robots Syst.}, 2013, pp.
  3697--3703.

\bibitem{stuckler2012semantic}
J.~St{\"u}ckler, N.~Biresev, and S.~Behnke, ``Semantic mapping using
  object-class segmentation of {RGB-D} images,'' in \emph{Proc. IEEE/RSJ Int.
  Conf. Intell. Robots Syst.}, 2012, pp. 3005--3010.

\bibitem{xiang2017darnn}
Y.~Xiang and D.~Fox, ``{DA-RNN}: Semantic mapping with data associated
  recurrent neural networks,'' in \emph{Robotics: Science and Systems}, 2017.

\bibitem{kim20133d}
B.-s. Kim, P.~Kohli, and S.~Savarese, ``{3D} scene understanding by
  voxel-{CRF},'' in \emph{Proc. IEEE Int. Conf. Computer Vision}, 2013, pp.
  1425--1432.

\bibitem{hermans2014dense}
A.~Hermans, G.~Floros, and B.~Leibe, ``Dense {3D} semantic mapping of indoor
  scenes from {RGB-D} images,'' in \emph{Proc. IEEE Int. Conf. Robot Automat.},
  2014, pp. 2631--2638.

\bibitem{zhao2016building}
Z.~Zhao and X.~Chen, ``Building {3D} semantic maps for mobile robots using
  {RGB-D} camera,'' \emph{Intell. Serv. Robot.}, vol.~9, no.~4, pp. 297--309,
  2016.

\bibitem{kundu2014joint}
A.~Kundu, Y.~Li, F.~Dellaert, F.~Li, and J.~M. Rehg, ``Joint semantic
  segmentation and {3D} reconstruction from monocular video,'' in
  \emph{European Conf. Computer Vision}.\hskip 1em plus 0.5em minus 0.4em\relax
  Springer, 2014, pp. 703--718.

\bibitem{Silberman:ECCV12}
P.~K. Nathan~Silberman, Derek~Hoiem and R.~Fergus, ``Indoor segmentation and
  support inference from {RGBD} images,'' in \emph{ECCV}, 2012.

\bibitem{tipping2004bayesian}
M.~E. Tipping, ``Bayesian inference: An introduction to principles and practice
  in machine learning,'' \emph{Lecture notes in computer science}, vol. 3176,
  pp. 41--62, 2004.

\bibitem{bishop2006pattern}
C.~M. Bishop, \emph{Pattern recognition and machine learning}.\hskip 1em plus
  0.5em minus 0.4em\relax springer, 2006.

\bibitem{scholkopf2002learning}
B.~Sch{\"o}lkopf and A.~J. Smola, \emph{Learning with kernels: support vector
  machines, regularization, optimization, and beyond}.\hskip 1em plus 0.5em
  minus 0.4em\relax MIT press, 2002.

\bibitem{fawcett2006introduction}
T.~Fawcett, ``An introduction to {ROC} analysis,'' \emph{Pattern recognition
  letters}, vol.~27, no.~8, pp. 861--874, 2006.

\bibitem{Geiger2010ACCV}
A.~Geiger, M.~Roser, and R.~Urtasun, ``Efficient large-scale stereo matching,''
  in \emph{Asian Conf. Computer Vision}, 2010.

\bibitem{segal2009generalized}
A.~Segal, D.~Haehnel, and S.~Thrun, ``Generalized-icp.'' in \emph{Robotics:
  Science and Systems}, vol.~2, no.~4, 2009.

\end{thebibliography}

\end{document}